\documentclass[letterpaper]{article}
\usepackage{aaai2026}
\usepackage{amssymb}
\usepackage{times}
\usepackage{helvet}
\usepackage{makecell}

\usepackage{courier}
\usepackage{natbib}
\usepackage{algorithmic}
\usepackage[hyphens]{url}                    
\usepackage{microtype}                       
\usepackage{amsmath,amssymb,amsfonts}
\usepackage{algorithmic}
\usepackage{url}
\usepackage{graphicx}
\usepackage{textcomp}
\usepackage{xcolor}
\usepackage{multirow}
\usepackage{booktabs}
\usepackage{dsfont}  
\usepackage{caption}
\usepackage{float}
\usepackage{etoolbox}
\AtBeginEnvironment{figure}{\setlength{\intextsep}{4pt plus 2pt minus 2pt}}
\makeatletter
\renewcommand\subsection{\@startsection{subsection}{2}{\z@}%
  {5pt plus 2pt minus 2pt}
  {2pt plus 1pt minus 1pt}
  {\normalfont\normalsize\bfseries}} 
\makeatother
\makeatletter
\def\maketitle{%
  \par%
  \begingroup
    \def\thefootnote{\fnsymbol{footnote}}%
    \twocolumn[\@maketitle] \@thanks%
  \endgroup%
  \setcounter{footnote}{0}%
  \let\maketitle\relax%
  \let\@maketitle\relax%
  \gdef\@thanks{}%
  \gdef\@author{}%
  \gdef\@title{}%
  \let\thanks\relax%
}
\makeatother

\begin{document}
 \title{MVRD-Bench: Multi-View Learning and Benchmarking for Dynamic Remote Photoplethysmography under Occlusion}
\author{
    Zuxian He\textsuperscript{1},
    Xu Cheng\textsuperscript{1},
    Zhaodong Sun\textsuperscript{1},
    Haoyu Chen\textsuperscript{2},
    Jingang Shi\textsuperscript{3},
    Xiaobai Li\textsuperscript{4},
    Guoying Zhao\textsuperscript{2} \\
    \textsuperscript{1}Nanjing University of Information Science and Technology \\
    \textsuperscript{2}University of Oulu \\
    \textsuperscript{3}Xi'an Jiaotong University \\
    \textsuperscript{4}Zhejiang University \\
}

\maketitle
\begin{abstract}
Remote photoplethysmography (rPPG) is a non-contact technique that estimates physiological signals by analyzing subtle skin color changes in facial videos. Existing rPPG methods often encounter performance degradation under facial motion and occlusion scenarios due to their reliance on static and single-view facial videos. Thus, this work focuses on tackling the motion-induced occlusion problem for rPPG measurement in unconstrained multi-view facial videos. Specifically, we introduce a \textbf{M}ulti-\textbf{V}iew \textbf{r}PPG \textbf{D}ataset (MVRD), a high-quality benchmark dataset featuring synchronized facial videos from three viewpoints under stationary, speaking, and head movement scenarios to better match real-world conditions. We also propose MVRD-rPPG, a unified multi-view rPPG learning framework that fuses complementary visual cues to maintain robust facial skin coverage, especially under motion conditions. Our method integrates an Adaptive Temporal Optical Compensation (ATOC) module for motion artifact suppression, a Rhythm-Visual Dual-Stream Network to disentangle rhythmic and appearance-related features, and a Multi-View Correlation-Aware Attention (MVCA) for adaptive view-wise signal aggregation. Furthermore, we introduce a Correlation Frequency Adversarial (CFA) learning strategy, which jointly enforces temporal accuracy, spectral consistency, and perceptual realism in the predicted signals. Extensive experiments and ablation studies on the MVRD dataset demonstrate the superiority of our approach. In the MVRD movement scenario, MVRD-rPPG achieves an MAE of 0.90 and a Pearson correlation coefficient (R) of 0.99. The source code and dataset will be made available.
\end{abstract}

\section{Introduction}
Remote photoplethysmography (rPPG) is of significance for healthcare~\cite{MTTS-CAN}, emotion analysis~\cite{MMSE-HR}. It aims to estimate physiological signals such as heart rate (HR), respiration frequency (RF), and heart rate variability (HRV) from facial videos in a non-contact manner. In recent years, rPPG has emerged as an innovative alternative for acquiring physiological signals, leveraging camera technology to detect subtle color fluctuations on human skin.

\begin{figure*}[t]
    \centering
    \includegraphics[width=0.7\linewidth]{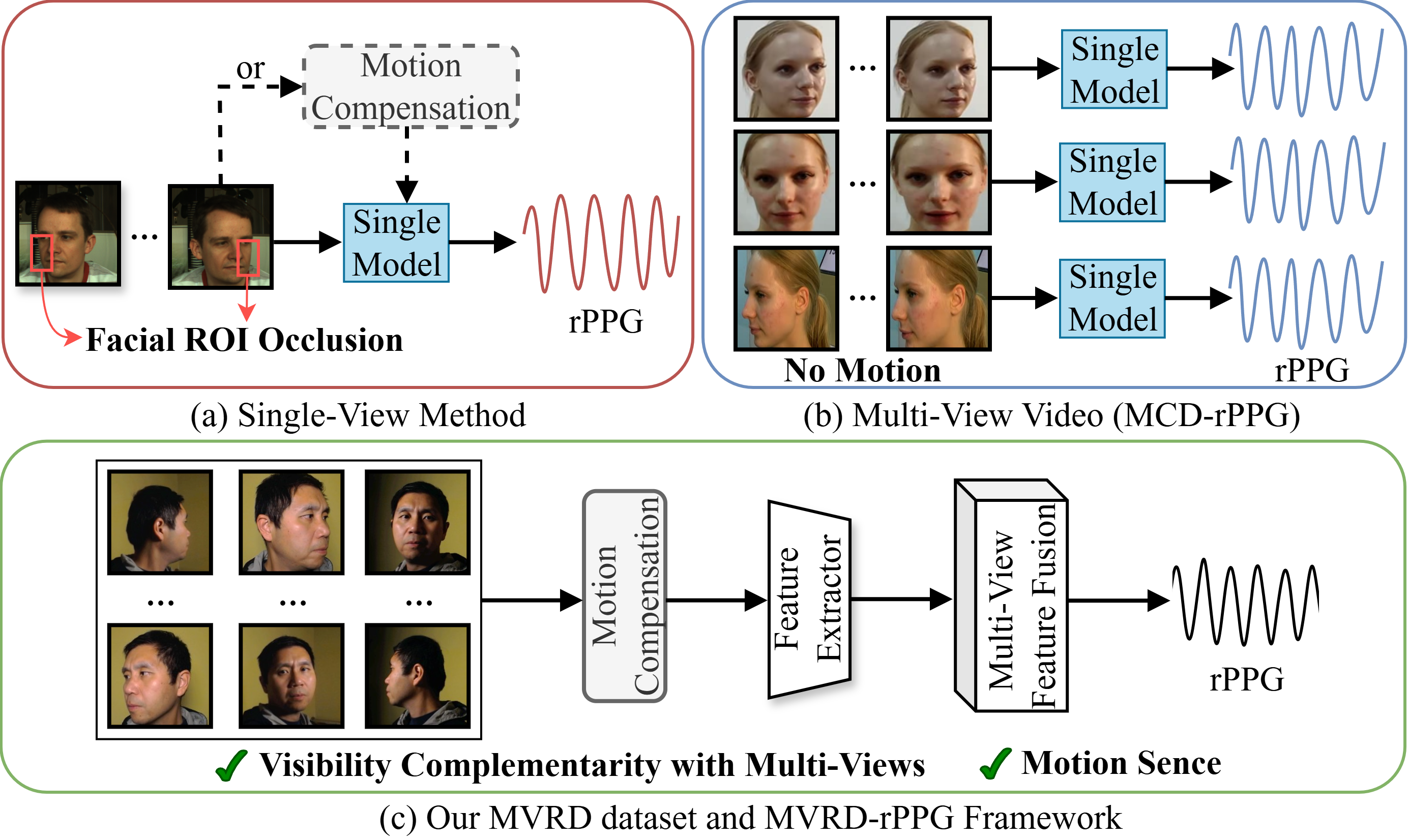}
\caption{
Motivation and challenges in the movement scenario. (a) Single-view rPPG remains vulnerable to motion-induced facial ROI occlusion even with motion compensation. (b) Existing multi-view data (e.g., MCD-rPPG) are captured independently in static settings and process views. (c) We construct MVRD dataset and develop a multi-view rPPG measurement framework for robust estimation in movement scenarios.}
\label{motivation}
\end{figure*}

Despite significant progress in rPPG~\cite{verkruysse,POS,ICA,deepphys,ST-networks,physformer,iBVPNet,contrast-phys,contrast-phys+,yue2023,SiNC-rPPG,mcd_rppg}, existing datasets and methods remain limited in dynamic scenarios, as illustrated in Fig.~\ref{motivation}. As shown in Fig.~\ref{motivation}(a), most existing rPPG methods operate on single-view facial videos. Under natural head movements, a single view can easily suffer from facial ROI occlusion and motion artifacts, which directly degrade the performance of rPPG estimation. Although motion compensation can reduce interference to some extent, it cannot solve the visibility loss caused by the single-view bottleneck. In Fig.~\ref{motivation}(b), Egorov et al.~\cite{mcd_rppg} introduced the first public multi-view MCD-rPPG dataset for analyzing view variations in rPPG. However, it is restricted to static scenarios and cannot capture realistic view changes caused by natural head movements. In addition, existing public multi-view datasets \cite{MHAD,mcd_rppg} and the corresponding rPPG methods did not establish a dynamic multi-view fusion prototype for rPPG estimation, which limits the use of complementary facial visibility across views during motion. As a result, dynamic multi-view rPPG under motion-induced occlusion remains underexplored. 

To address these limitations, we focus on dynamic multi-view rPPG estimation, as shown in Fig.~\ref{motivation}(c). We construct a Multi-View rPPG Dataset (MVRD) with synchronized recordings from three viewpoints under stationary, speaking, and head movement scenarios, and further develop a unified multi-view rPPG learning framework that combines motion compensation with cross-view fusion for robust rPPG estimation under motion-induced occlusion.

Generally, the contributions of this work can be summarized as follows:
\begin{itemize}
    \item
    We construct a Multi-View Remote Photoplethysmography dataset (MVRD), which contains facial recordings from three viewpoints in stationary, speaking, and head movement scenarios to study the motion robustness under the multi-view setting. 
    \item
    We propose the first unified multi-view rPPG learning framework (MVRD-rPPG) utilizing visual cues of different views to maintain full facial skin coverage and improve motion robustness, encompassing three carefully designed components: Adaptive Temporal Optical Compensation module, Rhythm-Visual Dual-Stream Network, and Multi-View Correlation-Aware Attention. 
    \item
    We develop a Correlation Frequency Adversarial (CFA) learning strategy to capture more stable power spectral density distributions and improve rPPG signal measurement in fused prediction, which jointly enforces temporal accuracy, spectral consistency, and perceptual realism in the predicted signals. 
    \item
    Extensive experiments and ablation studies on MVRD and public datasets demonstrate the robustness and generalization ability of the proposed framework in both dynamic multi-view setting and common single-view setting.
\end{itemize}

\section{Related Work}
\noindent\textbf{Video-Based Remote Physiological Measurement.} Extracting rPPG signals from facial videos is challenging due to their subtle nature. Early signal processing methods relied on priors. Verkruysse~\cite{verkruysse} first demonstrated the feasibility of extracting physiological signals under ambient light. PCA and ICA~\cite{ICA} decompose temporal signals, while CHROM~\cite{CHROM} and POS~\cite{POS} project skin-color signals into subspaces to estimate HR using skin boundary and mask filters. 

With the development of deep learning, supervised-based rPPG models have been proposed to learn spatial-temporal features from raw frames and ground-truth signals. DeepPhys~\cite{deepphys} introduces convolutional attention to disentangle appearance and motion. PhysNet ~\cite{ST-networks} models temporal dynamics with 3D CNNs. MTTS-CAN~\cite{TS-CAN} performs multi-task learning for pulse and attention, while TS-CAN~\cite{TS-CAN,TS-CAN+} improves temporal consistency via temporal shifts. However, these models are dependent on high-quality synchronized labels.

Unsupervised and self-supervised rPPG methods have emerged to address this limitation. AutoHR~\cite{Autohr} uses a spatial-temporal autoencoder for signal reconstruction. SiNC~\cite{SiNC-rPPG} introduces frequency-domain consistency constraints. Contrast-Phys~\cite{contrast-phys,contrast-phys+} employs a contrastive framework to pull together rPPG signals from the same video and separate those from different videos. Yue et al.~\cite{yue2023} propose a frequency-inspired self-supervised learning strategy that learns rPPG signals from unlabeled facial videos via contrastive frequency constraints. These methods work well under stable frontal views but often degrade under occlusion or large head movement.

\noindent\textbf{Optical Flow for Motion Compensation.}
Head movements and facial expressions introduce non-rigid motion artifacts that degrade the rPPG signal quality. Optical flow has been utilized to estimate the pixel-level displacement between frames to reduce motion-related distortions. DeepPhys~\cite{deepphys} uses frame difference maps to guide attention learning.  Gong et al.~\cite{FFEN} proposed a flow-guided attention mechanism to emphasize temporally consistent regions. These studies reveal that motion-aware representations improve robustness in the context of moderate movement. However, performance drops significantly under large head movements. Li et al.~\cite{PFE-TFA} utilized optical flow-based temporal alignment to suppress motion-induced noise in rPPG signals, enhancing robustness under head movement.

\noindent\textbf{Public Datasets for rPPG.} Benchmark datasets have been crucial in advancing the development of rPPG models by providing standardized evaluation frameworks. Early datasets, such as UBFC-rPPG~\cite{UBFC-rPPG} and OBF~\cite{obf}, were collected under conditions of small movement and consistent illumination, which limited their generalizability. The PURE ~\cite{PURE} introduced moderate head motions but remained a single-view setup, which was less robust to head movements in terms of data sources. The MMSE-HR~\cite{MMSE-HR} and VIPL-HR~\cite{VIPL-HR} expanded the diversity of data on lighting conditions and head poses, making them more applicable to real-world scenarios. However, these datasets still use a single camera and lack mechanisms to address face occlusions caused by natural head movements. 

Recent work by Egorov et al.~\cite{mcd_rppg} introduced the first publicly available multi-view dataset (MCD-rPPG) for rPPG-related analysis, which contains a large number of subjects and multi-camera recordings. However, it is restricted to static scenarios, failing to model natural head movements and face occlusions. More importantly, MCD-rPPG only tested previous rPPG methods on single-view videos and did not investigate multi-view fusion strategies. Therefore, we construct a high-quality Multi-View RPPG Dataset (MVRD), which contains 369 synchronized video sequences from 41 participants. MVRD dataset simultaneously records facial videos from three viewpoints (left, center, and right) under stationary, speaking, and natural head movement scenarios, ensuring unobstructed facial views for motion robustness.

\section{The Multi-View rPPG Dataset}
\begin{figure*}[t]
\centering
\includegraphics[width=0.8\linewidth]{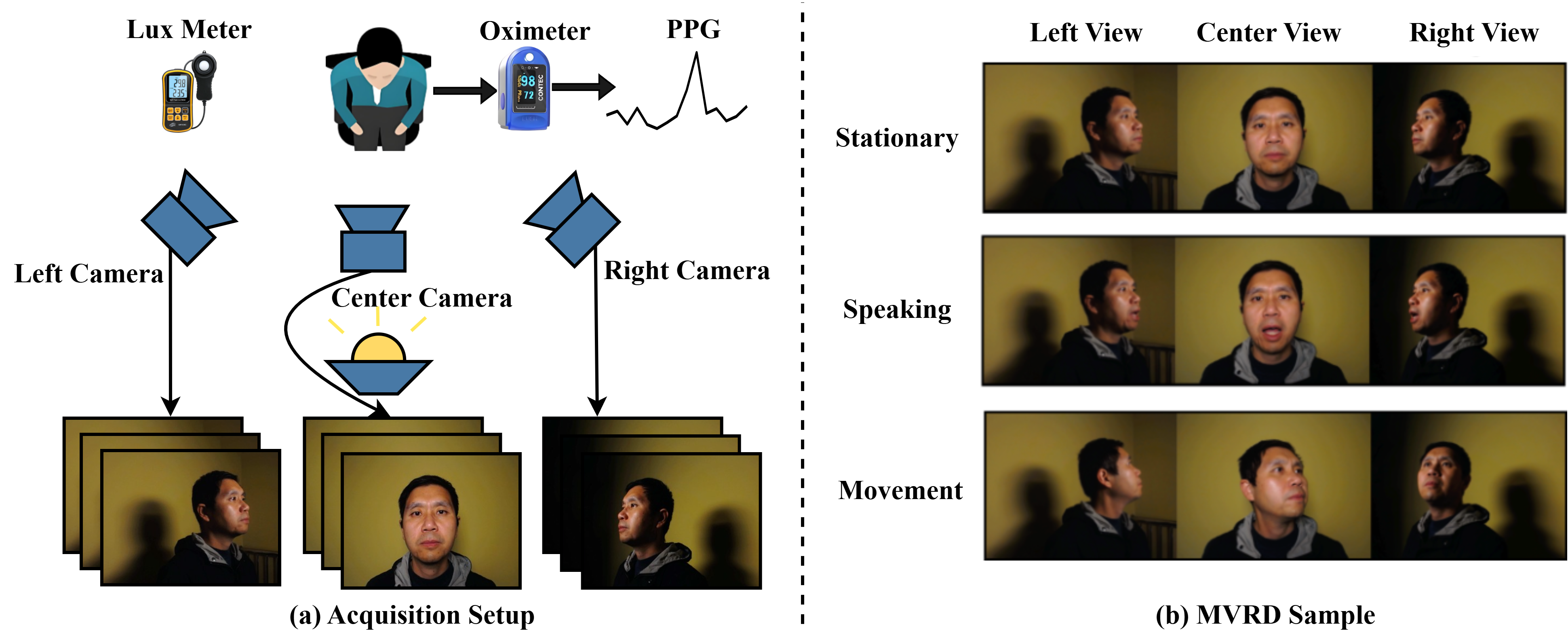}
    \caption{Overview of the MVRD acquisition setup. (a) Data collection scenario and recording devices. (b) An example subject from MVRD under the three scenarios.}
    \label{scene}
\end{figure*}
To support dynamic multi-view rPPG under motion-induced occlusion, we construct a high-quality Multi-View rPPG Dataset (MVRD) with synchronized facial videos from three spatially distributed viewpoints and aligned ground-truth physiological signals, as shown in Fig.~\ref{scene}.

\subsection{Multi-View Capture under Motion}
The MVRD dataset is recorded in a closed room with three Logitech C930E RGB cameras positioned at approximately $-45^\circ$, $0^\circ$, and $+45^\circ$ around the subject, each at a fixed distance of one meter. The ambient lighting is fixed at 35 Lux using LED light sources placed in front of the subject. This configuration ensures the full facial skin coverage, even during natural head movements.

MVRD dataset comprises 369 multi-view videos with corresponding ground-truth physiological signals, obtained from 41 subjects. Each subject is recorded in three sessions: stationary, speaking, and head movement. During each session, three synchronized video streams are captured at 30 fps for 65 seconds with $480\times640$ resolution. All videos are temporally synchronized with ground-truth PPG signals recorded at 60 Hz using a CONTEC CMS50E fingertip pulse oximeter. Raw oximeter signals are resampled to ensure temporal alignment between physiological signals and video frames. All participants provided informed consent for data collection and usage. More details are provided in the supplementary material.

\begin{figure*}[t]
\centering
        \includegraphics[width=1\linewidth]{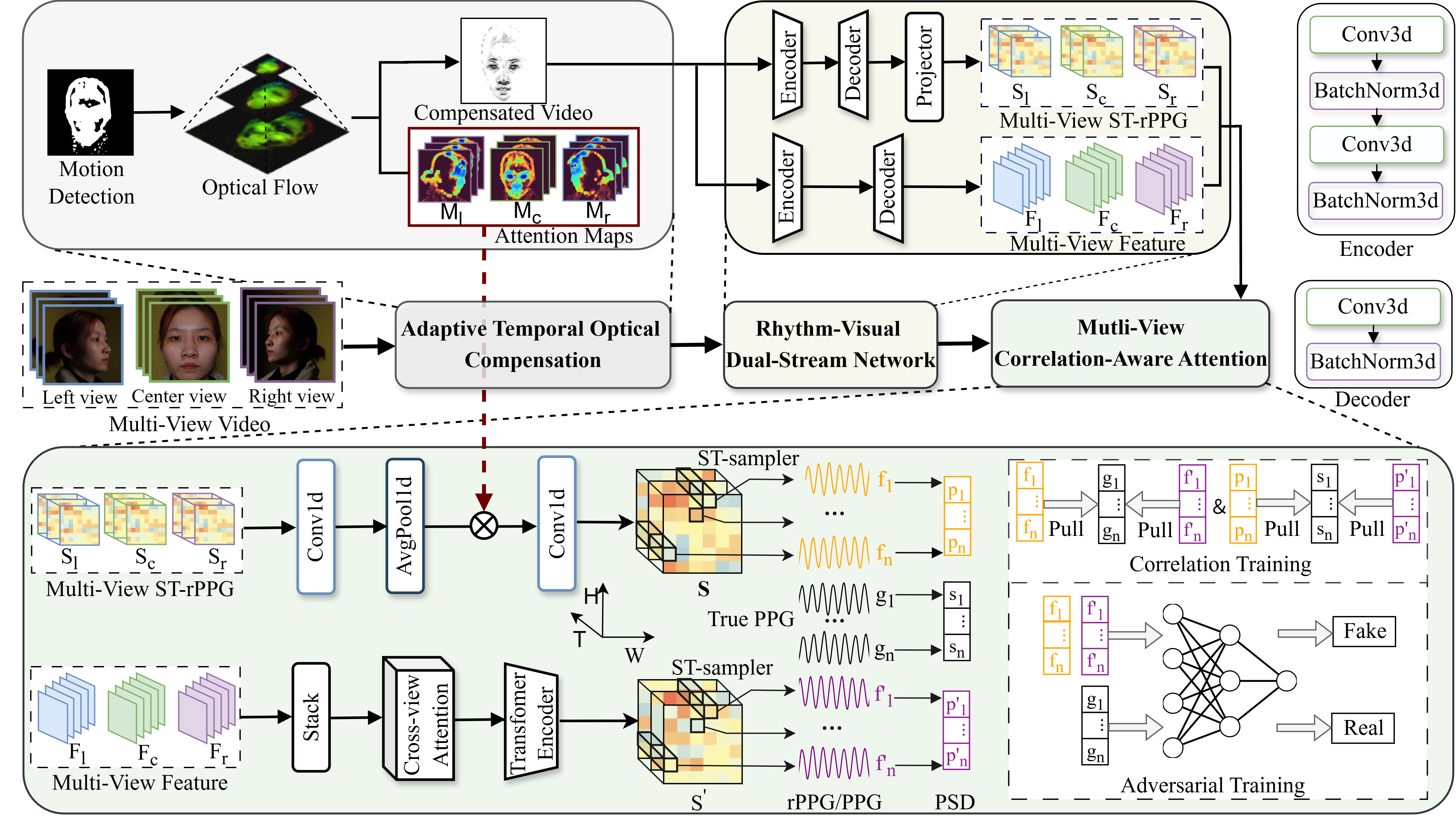}
    \caption{Architecture of the proposed MVRD-rPPG framework. Three synchronized video streams undergo (1) per-view Adaptive Temporal Optical Compensation to suppress motion artifacts, (2) Rhythm-Visual Dual-Stream Network to capture complementary spatial-temporal and perceptual cues, and (3) a three-stage Multi-View Correlation-Aware Attention comprises flow-noise-aware ST-rPPG aggregation, cross-view temporal attention, and gated synergy fusion.}
    \label{method}
\end{figure*}
\section{Methodology}
In this section, we introduce MVRD-rPPG, a novel framework that processes synchronized video streams from three spatially distributed cameras and fuses different views to produce robust rPPG signals. The overall pipeline of the proposed method is shown in Fig.~\ref{method}.

\subsection{Adaptive Temporal Optical Compensation}
\begin{figure}[t]
    \centering
    \includegraphics[width=0.9\linewidth]{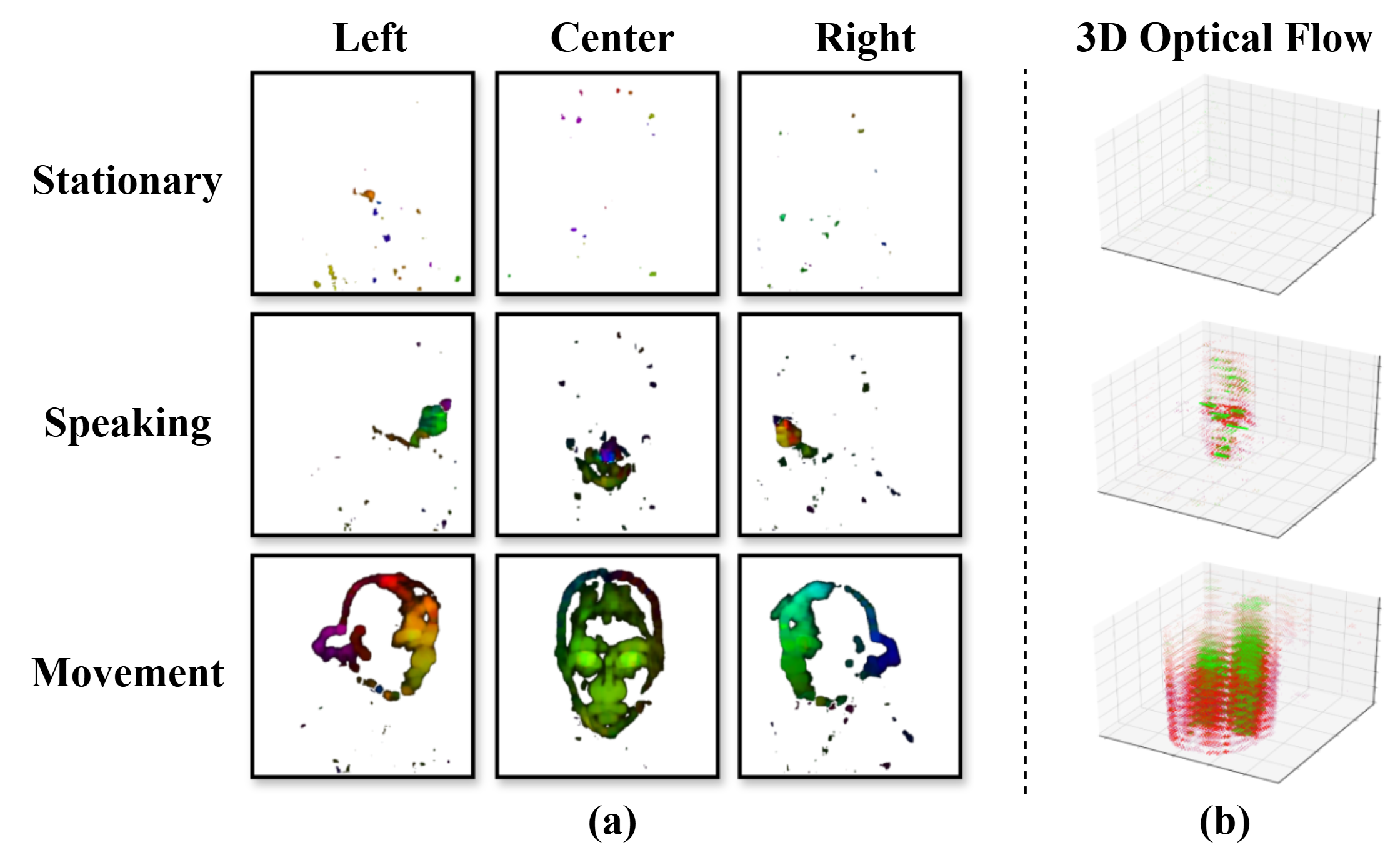}
    \caption{
    Visualization of 2D and 3D optical flow distributions across different scenarios on the MVRD dataset. (a) The 2D dense optical flow distributions in the three
    scenarios. (b) Aggregated 3D optical flow field from three synchronized camera views.
    }
    \label{opticalflow}
\end{figure}

Head movements introduce additional color variations that distort subtle pulsatile signals and hinder reliable rPPG extraction. Fig.~\ref{opticalflow}(a) shows the HSV-encoded 2D optical flow maps in stationary, speaking, and movement scenarios. Motion is negligible in the stationary scene, and the optical flow distribution is almost zero. There is a widespread non-rigid facial motion in the movement scene, which seriously degrades signal quality.  Fig.~\ref{opticalflow}(b) shows 3D optical flow fields that our strategy can aggregate from three views when motion occurs. 

To address this issue, we introduce an Adaptive Temporal Optical Compensation (ATOC) strategy inspired by first-order motion modeling~\cite{first-oder-motion}. ATOC involves three steps: motion mask generation, confidence-aware optical flow estimation, and selective pixel-wise warping. It is applied only to the movement subset of MVRD, ensuring compensation in dynamic scenes without affecting others. 
Specifically, let $I_v^t(x, y)$ denote the RGB intensity at pixel $(x, y)$ in frame $t$ from view $v \in \{l, c, r\}$, corresponding to the left, center, and right views. To localize motion, we first apply background subtraction~\cite{IAGMM} between adjacent frames to obtain a binary motion mask:
\begin{equation}
    M_v^t(x, y) = 
    \begin{cases}
      1, & \bigl|I_v^{t+1}(x, y) - I_v^t(x, y)\bigr| > \delta_m, \\
      0, & \text{otherwise}.
    \end{cases}
    \label{M_v}
\end{equation}
We further refine $M_v^t$ using morphological closing and dilation to enhance spatial continuity, where $\delta_m$ is adaptively determined via background subtraction with automatic thresholding.

We model local motion near salient regions by estimating affine parameters. The keypoints $\{\mathbf{p}_k\}_{k=1}^K$ are extracted using the OpenFace toolkit~\cite{OpenFace}, which provides robust facial landmark detection. For each keypoint $\mathbf{p}_k$, the local motion is approximated by:
\begin{equation}
    \mathbf{T}_v^t(\mathbf{z}) \approx \mathbf{A}_k (\mathbf{z} - \mathbf{p}_k) + \mathbf{b}_k,
    \label{eq:affine}
\end{equation}
where $\mathbf{z} = (x, y)^\top$ is a pixel location, and $\mathbf{A}_k \in \mathbb{R}^{2 \times 2}$ and $\mathbf{b}_k \in \mathbb{R}^2$ are affine parameters estimated from the frame-to-frame displacements of facial landmarks, yielding a local linear approximation of motion around each keypoint. This affine model implicitly represents local motion, capturing non-rigid facial dynamics faster than traditional optical flow.

High-confidence motion regions are selected by combining the motion mask with the magnitude of affine displacement:
\begin{equation}
    \Omega_v^t = \{(x, y) \mid \|\mathbf{T}_v^t(\mathbf{z})\| > \tau, (x, y) \in M_v^t\}.
    \label{Omega_v}
\end{equation}
The threshold $\tau$ is empirically determined and kept fixed across all experiments. Each pixel in $\Omega_v^t$ is backward-warped from the previous frame using the affine model to yield a compensated image:
\begin{equation}
    \hat{I}_v^t(x, y) = I_v^{t-1}\big(x - \Delta x, y - \Delta y\big),
    \label{I_t}
\end{equation}
where $(\Delta x, \Delta y) = \mathbf{T}_v^t(\mathbf{z}) $. Pixels outside $\Omega_v^t$ remain unchanged. The compensated sequence $\{\hat{I}_v^t\}_{t=1}^T$ is then fed into the Rhythm-Visual Dual-Stream Network for feature extraction. By preserving temporal alignment and suppressing non-rigid motion artifacts, ATOC enhances physiological signal stability.

\subsection{Rhythm-Visual Dual-Stream Network}
Robust rPPG estimation under head motion and occlusion requires modeling two complementary cues. The pulse signal is manifested as structured spatiotemporal rhythm patterns distributed over facial regions, while reliable estimation also depends on preserving local appearance evidence such as skin texture, reflectance variations, and view-dependent photoplethysmographic cues. In a single-stream architecture, these cues are entangled in a shared representation, making the learned features more susceptible to motion interference, pose changes, and partial occlusion. To explicitly decouple them, we propose a rhythm-visual dual-stream network consisting of a Rhythm-Structural Stream and a Visual-Perceptual Stream.

Given motion-compensated video sequences $\{\hat I_v^t\}_{t=1}^T$ from each view $v \in \{l,c,r\}$, both streams process the same input in parallel. The Rhythm-Structural Stream emphasizes pulse-related spatiotemporal dependencies, whereas the Visual-Perceptual Stream preserves appearance-sensitive evidence that may be weakened when learning rhythm-dominant features alone.

\noindent\textbf{Rhythm-Structural Stream.}
This stream is designed to model structured spatial-temporal patterns associated with physiological pulse propagation and spatially correlated skin-color dynamics. It adopts a 3D CNN encoder-decoder architecture consisting of an initial convolution block, hierarchical temporal-spatial downsampling layers, a mid-level feature transformation block, and temporal upsampling layers to restore temporal resolution. A spatial-temporal projection head is then applied to aggregate the decoded features by spatial pooling opepration followed by a 3D convolution, producing structured spatial-temporal rPPG representations (ST-rPPG), \(
\mathbf{S}_v=\{\mathbf{S}_l,\mathbf{S}_c,\mathbf{S}_r\}\in\mathbb{R}^{B\times 3\times N\times T},\)
where $B$ is the batch size, $T$ is the temporal length, and $N=P^2$ denotes the number of spatial tokens obtained by dividing the spatial map into $P\times P$ patches with $P=2$. By enforcing a compact tokenized representation over space and time, this stream focuses on global and long-range rhythm structure rather than local appearance fluctuations.

\noindent\textbf{Visual-Perceptual Stream.}
While rhythm-oriented representations are beneficial for capturing pulse dynamics, they may suppress subtle local cues that remain informative under view changes and partial occlusion. Therefore, we introduce a second stream to preserve appearance-sensitive information relevant to rPPG estimation, including local texture variations, fine-grained reflectance changes, and view-dependent photoplethysmographic clues. This stream uses the same 3D CNN encoder-decoder backbone for architectural consistency, but learns a complementary representation \(\mathbf{F}_v=\{\mathbf{F}_l,\mathbf{F}_c,\mathbf{F}_r\}\in\mathbb{R}^{B\times 3\times D\times T},\)
where $D$ denotes the feature dimension.

The two streams are complementary. The Rhythm-Structural Stream provides motion-robust global physiological structure, while the Visual-Perceptual Stream preserves local appearance evidence that may still be reliable in challenging views. Their outputs are subsequently aligned and fed into the multi-view correlation-aware attention module for cross-view interaction and adaptive fusion.

\subsection{Multi-View Correlation-Aware Attention.}
We propose a Multi-View Correlation-Aware Attention (MVCA) to fuse ST-rPPG signals $\mathbf{S}_v $ and multi-view Features $\mathbf{F}_v $, which consists of flow-noise-aware ST-rPPG Aggregation, cross-view temporal attention, and gated synergy fusion to improve the precision of rPPG signal prediction~\cite{FFEN,M3net}. MVCA facilitates complementary feature fusion across three synchronized views, ensuring coverage of facial skin regions relevant to rPPG. Unlike single-view settings that are prone to occlusion during large head movements, MVRD promotes more reliable ST-rPPG signal reconstruction in multi-view movement scenes.

\noindent\textbf{Flow-Noise-Aware ST-rPPG Aggregation.} On the MVRD dataset, natural head movements introduce motion noise in each view, causing motion-induced artifacts that corrupt skin pixel dynamics and degrade rPPG signal quality. To suppress motion-corrupted views, we propose a flow-noise-aware ST-rPPG Aggregation strategy to weight the importance of ST-rPPG signals $\mathbf{S}_v$, which computes the flow-noise score of each view by spatially averaging the flow magnitude. The score is defined as:
\begin{equation}
n_v = \frac{1}{|M_v^t|} \sum_{(x, y) \in M_v^t} \|\mathbf{T}_v^t(\mathbf{z})\|, v \in \{l,c,r\}
\end{equation}
where \( n_v \) represents the flow-noise score in the view $v$; \( M_v^t \) is the motion mask. A higher flow-noise score leads to lower importance of the ST-rPPG signals.

Then, the flow-noise weight $w_v$ from the view $v$ is derived as follows.
\begin{equation}
w_v = \frac{1 / (n_v + \epsilon)}{\sum_{i \in \{l, c, r\}} 1 / (n_i + \epsilon)},
\end{equation}
where \(\epsilon \) is a small constant to avoid division by zero. 

Further, the flow-noise weight is utilized to aggregate the ST-rPPG signals from each view, producing an aggregated ST-rPPG block as follows.
\begin{equation}
\begin{aligned}
\mathbf{S} = \mathrm{Conv1D}\Bigl(\sum_{v \in \{l, c, r\}} w_v \mathbf{S}_v \Bigr),
\end{aligned}
\label{S}
\end{equation}
where $\mathbf{S} \in \mathbb{R}^{B \times N \times D}$ is the aggregated ST-rPPG signals of three views.

\noindent\textbf{Cross-View Temporal Attention.}
We propose a cross-view temporal attention module to align and aggregate appearance features $\mathbf{F}_v$ from different views, producing a holistic spatiotemporal representation for physiological signal estimation~\cite{M3net}.

Specifically, the appearance features from the left, center, and right views are stacked as
\begin{equation}
\mathbf{F} = \mathrm{Stack}(\mathbf{F}_l, \mathbf{F}_c, \mathbf{F}_r),
\label{F}
\end{equation}
where $\mathbf{F}\in\mathbb{R}^{B\times 3\times D\times T}$, with $B$, $D$, and $T$ denoting the batch size, feature dimension, and temporal length, respectively. For each time step $t$, we denote the cross-view feature slice by $\mathbf{F}_t\in\mathbb{R}^{B\times 3\times D}$. To model inter-view correlations at each time step, we project $\mathbf{F}_t$ into query, key, and value embeddings using three learnable projection matrices $\mathbf{W}_Q,\mathbf{W}_K,\mathbf{W}_V\in\mathbb{R}^{D\times D}$, which can be written as:
\begin{equation}
\mathbf{Q} = \mathbf{W}_Q \mathbf{F_t}, \mathbf{K} = \mathbf{W}_K \mathbf{F_t}, \mathbf{V} = \mathbf{W}_V \mathbf{F_t}.
\end{equation}
The enhanced cross-view feature at time step $t$ is then computed by
\begin{equation}
\mathbf{F}'=
\mathrm{softmax}\Bigl(\frac{\mathbf{Q}\mathbf{K}^\top}{\sqrt{D}}\Bigr)\mathbf{V},
\label{F_prime}
\end{equation}
where $\mathbf{F}'\in\mathbb{R}^{B\times 3\times D}$ denotes the attention-enhanced feature across the three views at time step $t$. By aggregating the view dimension over all time steps, we obtain a fused temporal feature sequence $\tilde{\mathbf{F}}\in\mathbb{R}^{B\times T\times D}$. The cross-view attention module captures inter-view correlations at each time step, while a Transformer encoder is further applied to $\tilde{\mathbf{F}}$ to model long-range temporal dependencies across frames. The temporally encoded sequence is then projected into an ST-rPPG representation with $N$ spatial tokens:
\begin{equation}
\mathbf{S}'=\mathrm{Linear}\bigl(\mathrm{Transformer}(\tilde{\mathbf{F}})\bigr),
\label{S_prime}
\end{equation}
where $\mathbf{S}'\in\mathbb{R}^{B\times N\times T}$ denotes the resulting cross-view aggregated ST-rPPG block.

\noindent\textbf{Gated Synergy Fusion Module.}
We propose a Gated Synergy Fusion module~\cite{Autohr} to adaptively integrate dual ST-rPPG features $\mathbf{S}$ and $\mathbf{S}'$ with learned dynamic weighting. The two features can be fused by:
\begin{equation}
    \mathbf{U} = \sigma(\beta)\,{\mathbf{S}} + (1 - \sigma(\beta))\,{\mathbf{S}}', \label{U}
\end{equation}
where $\sigma(\cdot)$ denotes the sigmoid function; $\beta$ denotes a learnable scalar gating parameter trained by MVCA, which allows the network to adaptively balance the contribution between $\mathbf{S}$ and ${\mathbf{S}}'$. $\beta$ is initialized to 0.5.

To learn the gating coefficient $\sigma(\beta)$, we utilize a structured spatiotemporal sampling strategy to align two features in spatial and temporal perspective. Specifically, the feature maps $\mathbf{S}$ and ${\mathbf{S}}'$ of each frame are respectively divided into $N =P \times P$ spatial regions, generating $N$ spatial patches along the spatial dimension. Further, each spatial patch extracts $K$ non-overlapping rPPG segments along the temporal dimension, yielding a total of $n = N \times K$ temporal segment pairs $\{\mathbf{f}_1, \dots, \mathbf{f}_n\}$ and $\{\mathbf{f}'_1, \dots, \mathbf{f}'_n\}$ from $\mathbf{S}$ and ${\mathbf{S}}'$, respectively. For the \emph{i}-th temporal segment pair $\{\mathbf{f}_i, \mathbf{f}'_i\}$, the corresponding ground-truth rPPG signal $\mathbf{g}_i$ is utilized to ensure consistency in time for model training. Each temporal segment pair is aligned to obtain triplet samples $\{\mathbf{f}_i, \mathbf{f}'_i, \mathbf{g}_i\}_{i=1}^{n}$.

Further, the fused feature $\mathbf{U}$ passes through a $1 \times 1$ Conv1D layer to generate feature $\mathbf{Y} \in \mathbb{R}^{B \times N \times T}$, whose dimension is the same as that of the ST-rPPG block.
\begin{equation}
    \mathbf{Y} = \mathrm{Conv1D}(\mathbf{U}) 
\label{Y}.
\end{equation}

\subsection{Correlation Frequency Adversarial Learning}
We train MVRD-rPPG end-to-end by using Correlation Frequency Adversarial (CFA) optimization strategy, comprising cross-view correlation-spectral consistency objectives, adversarial signal realism, and final signal fusion.

\noindent\textbf{Cross-View Correlation-Spectral Consistency Loss.} To ensure temporal and spectral consistency between the predicted and ground-truth rPPG signals, we develop a general loss from the correlation and frequency domain aspects.

First, we employ the Pearson correlation loss to measure the linear similarity between the predicted signals $\mathbf{f}_i, \mathbf{f}'_i$ and the ground-truth $\mathbf{g}_i$, defined as:
\begin{equation}
\mathcal{L}_{\text{Pearson}} = - \frac{1}{2n}\sum_{i=1}^n \sum_{\mathbf{f}\in\{\mathbf{f_i}, \mathbf{f_i'}\}} \frac{\mathrm{Cov}(\mathbf{f}, \mathbf{g}_i)}{\sqrt{\mathrm{Var}(\mathbf{f})\, \mathrm{Var}(\mathbf{g}_i)} + \epsilon},
\end{equation}
where $\mathrm{Cov}(\cdot)$ and $\mathrm{Var}(\cdot)$ denote covariance and variance operations, respectively.

Specifically, all triplet samples $\{ \mathbf{f}_i, \mathbf{f}'_i, \mathbf{g}_i \}_{i=1}^{n}$ are transformed and normalized using FFT. Then it passes through a bandpass filter with frequencies ranging from 0.7 to 4.0 Hz to suppress baseline drift and high-frequency noise, while preserving the key spectral components corresponding to physiological heart rate. 

Further, we present a Dual-Branch Power Spectral Density (PSD) consistency loss to enforce that the predicted frequency distributions are consistent with the ground-truth spectra signal in the heart rate band, which encourages the fused features to preserve physiologically plausible frequency characteristics. The triplet PSD samples are $\{\mathbf{p}_i, \mathbf{p}'_i, \mathbf{s}_i\}_{i=1}^{n}$. The PSD consistency loss is defined as: 
\begin{equation}
\mathcal{L}_{\text{PSD}} = \frac{\sum_{i=1}^{n} \left( \| \mathbf{p}_i - \mathbf{s}_i \|^2 + \| \mathbf{p}'_i - \mathbf{s}_i \|^2 \right)}{2n}.
\label{PSD}
\end{equation}

The proposed Correlation-Frequency strategy jointly improves both temporal and spectral accuracy.

\noindent\textbf{Adversarial Signal Realism Loss.} To enhance the perceptual authenticity and pulse waveform continuity of predicted signals, we utilize a 1D PatchGAN discriminator~\cite{pix2pix} and treat the MVCA-based prediction branch as the generator, encouraging it to produce realistic rPPG-like waveforms. The discriminator is trained to distinguish ground-truth rPPG signals $\mathbf{g}_i$ from predicted signals $\mathbf{f}$, which is written as:
\begin{equation}
\mathcal{L}_D = \frac{1}{2} \mathbb{E} \left[ (\mathrm{D}(\mathbf{g}_i) - 1)^2 + (\mathrm{D}(\mathbf{f}))^2 \right].
\end{equation}

The generator fools the discriminator by minimizing:
\begin{equation}
\mathcal{L}_{G} = \mathbb{E} \left[ \left( D(\mathbf{f}) - 1 \right)^2 \right].
\end{equation}


\noindent\textbf{Total Loss.} 
The overall loss function for generator training can be summarized as follows.
\begin{equation}
\mathcal{L}_{\text{total}} = \mathcal{L}_{\text{Pearson}} + \lambda_{\text{PSD}} \, \mathcal{L}_{\text{PSD}} + \lambda_G \, \mathcal{L}_{G}, 
\end{equation}
where $\lambda_{\text{PSD}}$ and $\lambda_G$ are utilized to balance temporal accuracy, spectral alignment, and perceptual realism, their values are empirically set and reported in the supplementary material.

\section{Experiments}
\begin{table*}[t]
\centering
\scriptsize  
\caption{Quantitative results on MVRD dataset for different scenarios (MAE↓, RMSE↓, R↑). The best and the second-best results are \textbf{bold}, respectively. All values are reported in beats per minute (bpm).}
\label{tab:movement}
\resizebox{\textwidth}{!}{
\scriptsize
\begin{tabular}{c|ccc|ccc|ccc|ccc|ccc}
\hline
\multirow{3}{*}{\centering Methods}
& \multicolumn{9}{c|}{Movement}
& \multicolumn{3}{c|}{Stationary}
& \multicolumn{3}{c}{Speaking} \\
\cline{2-10}\cline{11-13}\cline{14-16}
& \multicolumn{3}{c|}{Left}
& \multicolumn{3}{c|}{Center}
& \multicolumn{3}{c|}{Right}
& \multicolumn{3}{c|}{Center}
& \multicolumn{3}{c}{Center} \\
\cline{2-16}
& MAE & RMSE & R & MAE & RMSE & R & MAE & RMSE & R & MAE & RMSE & R & MAE & RMSE & R \\
\hline

ICA & 28.30 & 30.34 & 0.11 & 26.50 & 29.26 & 0.16 & 29.07 & 31.03 & 0.02 & 1.20 & 2.05 & 0.98 & 10.72 & 17.84 & 0.32 \\
CHROM & 20.75 & 24.69 & 0.10 & 17.75 & 21.97 & 0.46 & 24.35 & 27.65 & 0.04 & 1.63 & 4.29 & 0.92 & 3.17 & 7.26 & 0.73 \\
POS & 20.49 & 25.12 & 0.17 & 17.15 & 22.41 & 0.21 & 18.86 & 23.26 & 0.15 & 1.29 & 2.13 & 0.98 & 2.14 & 5.52 & 0.83 \\
PhysNet & 8.43 & 11.23 & 0.46 & 8.14 & 10.10 & 0.55 & 8.17 & 10.56 & 0.51 & 5.12 & 7.56 & 0.68 & 6.36 & 10.07 & 0.44 \\
PhysFormer & 6.23 & 7.82 & \textbf{0.65} & 5.95 & 7.12 & 0.74 & 6.34 & 8.02 & 0.61 & 4.52 & 5.50 & 0.77 & 5.29 & 7.95 & 0.73 \\
iBVPNet & 8.50 & 9.78 & 0.21 & 5.51 & 9.11 & 0.53 & 8.11 & 9.12 & 0.32 & 4.45 & 5.24 & 0.52 & 5.04 & 9.48 & 0.63 \\
Contrast-Phys+ & 7.76 & 9.07 & 0.56 & 6.23 & 7.67 & 0.69 & 7.31 & 8.21 & 0.52 & 0.89 & 1.04 & 0.99 & 2.02 & 3.80 & 0.89 \\
Yue2023 & 7.54 & 7.96 & 0.58 & 7.18 & 8.70 & 0.64 & 7.14 & 8.70 & 0.56 &0.73 & \textbf{1.01} & 0.99 & \textbf{1.65} & 2.56 & 0.93 \\
SiNC & 5.89 & 7.11 & 0.63 & 4.57 & 5.72 & 0.89 & 5.27 & 6.90 & 0.64 & 0.79 & 1.13 & 0.98 & 1.89 & \textbf{2.52} & \textbf{0.94}\\
Our Method & \textbf{3.99} & \textbf{5.98} & 0.63 & \textbf{1.10} & \textbf{1.22} & \textbf{0.97} & \textbf{3.20} & \textbf{4.75} & \textbf{0.65} & \textbf{0.89} & 1.05 & \textbf{0.99} & 2.00 & 3.56 & 0.90 \\
\hline
\end{tabular}
}
\end{table*}

\begin{table}[t]
\centering
\scriptsize  
\caption{Performance under different view availability conditions in the MVRD Movement scenario. \checkmark\ indicates the view is available.}
\label{tab:view-missing}
\begin{tabular}{cccccc}
\toprule
Left & Center & Right & MAE & RMSE & R \\
\midrule
\checkmark & \checkmark & \checkmark & \textbf{0.90} & \textbf{1.02} & \textbf{0.99} \\
~          & \checkmark & \checkmark & 1.05 & 1.03 & 0.96 \\
\checkmark & \checkmark & ~          & 0.99 & 1.02 & 0.98 \\
\checkmark & ~          & \checkmark & 3.07 & 3.97 & 0.67 \\
\checkmark & ~          & ~          & 3.99 & 5.98 & 0.63 \\
~          & \checkmark & ~          & 1.10 & 1.22 & 0.97 \\
~          & ~          & \checkmark & 3.20 & 4.75 & 0.65 \\
\bottomrule
\end{tabular}
\end{table}

We conduct comprehensive experiments to validate the effectiveness and robustness of the proposed method for rPPG signal estimation, which remains challenging under real-world conditions due to motion artifacts and occlusion. Evaluations are performed on our MVRD, including stationary, speaking, and movement scenarios. In addition, we carry out experiments to verify cross-dataset generalization on two public single-view datasets: PURE~\cite{PURE} and UBFC-rPPG~\cite{UBFC-rPPG}. Our MVRD-rPPG method is compared with handcrafted methods~\cite{verkruysse,CHROM,POS}, supervised rPPG methods~\cite{ST-networks,physformer,iBVPNet}, and unsupervised rPPG methods~\cite{contrast-phys+,yue2023,SiNC-rPPG}. More experimental details can be found in the supplementary materials.

\subsection{Experimental Setup}
\noindent\textbf{Datasets.} We construct MVRD, capturing synchronized facial videos from three viewpoints. The MVRD dataset includes recordings in stationary, speaking, and movement scenarios, enabling multi-view analysis of rPPG signals. Additionally, three public datasets are used to evaluate cross-dataset generalization, including the single-view PURE~\cite{PURE}, UBFC-rPPG~\cite{UBFC-rPPG}, and the multi-view MCD-rPPG~\cite{mcd_rppg}.

\noindent\textbf{Evaluation Metrics.} We utilize mean absolute error (MAE), root mean square error (RMSE), and Pearson correlation coefficient (R) between ground truth heart rates and predicted heart rates to evaluate the model performance.

\noindent\textbf{Data Preprocessing.} Raw PPG signals are resampled to 30 Hz to align with the video frame rate. Face bounding boxes are detected and resized to $128 \times 128$ pixels. Video-PPG pairs are stored in HDF5 format to enable data loading.

\noindent\textbf{Implementation Details.} Models are implemented in PyTorch and trained on NVIDIA Tesla V100 GPUs. Baselines include: (i) traditional hand-crafted methods ICA \cite{ICA}, CHROM \cite{CHROM}, POS \cite{POS}; (ii) supervised learning based rPPG methods PhysNet \cite{ST-networks}, PhysFormer \cite{physformer}, iBVPNet \cite{iBVPNet};
(iii) unsupervised learning based rPPG methods Contrast-Phys+ \cite{contrast-phys+}, yue2023\cite{yue2023}, SiNC\cite{SiNC-rPPG}. The Single-view model is trained independently for each scenario and view with a learning rate $1 \times 10^{-5}$ and a batch size of 4, lasting for 30 epochs. Our multi-view backbone model is initialized with the weights from the best single-view PhysNet to improve the single-view representation. Subsequently, we re-trained the model end-to-end within 20 epochs with a learning rate $1 \times 10^{-4}$. The input covers all three synchronized views, and we use 300 frame clips during both training and inference to maintain a consistent temporal setting while providing sufficient temporal context for robust rPPG estimation. For each scenario, the first 80\% of the samples are used for training, and the remaining 20\% are used for testing. More details are provided in the supplementary material.

\begin{table*}[t]
\centering
\scriptsize  
\caption{Cross-dataset evaluation on public rPPG datasets. All models are trained on MVRD Movement and tested without fine-tuning.}
\label{tab:cross_dataset}
\begin{tabular}{lccccccccc}
\toprule
\multirow{2}{*}{Method} & \multicolumn{3}{c}{PURE} & \multicolumn{3}{c}{UBFC-rPPG}& \multicolumn{3}{c}{MCD-rPPG} \\
\cmidrule(lr){2-4} \cmidrule(lr){5-7}\cmidrule(lr){8-10}
 & MAE & RMSE & R & MAE & RMSE & R & MAE & RMSE & R\\
\midrule
CHROM & 2.07 & 9.92 & 0.99 & 2.37 & 4.91 & 0.89 & - & - & - \\
PhysNet & 10.78 & 12.98 & 0.42 & 7.14 & 9.42 & 0.67 & 7.86 & 8.81 & 0.65\\
PhysFormer & 9.32 & 12.56 & 0.57 & 7.89 & 9.07 & 0.70 & 7.12 & 8.03 & 0.72\\
Contrast-Phys+ & 8.37 & 9.13 & 0.61 & 7.15 & 8.99 & 0.55 & 8.14 & 9.45 & 0.68 \\
SiNC & 3.19 & 5.13 & 0.79 & 1.95 & 3.12 & 0.85 & 2.69 & 3.28 & 0.91 \\
\textbf{Our Method} & \textbf{0.77} & \textbf{1.00} & \textbf{0.99} & \textbf{0.83} & \textbf{1.73} & \textbf{0.98} & \textbf{0.69} & \textbf{0.81} & \textbf{0.99} \\
\bottomrule
\end{tabular}
\end{table*}

\subsection{Experimental Results}
The following experiments demonstrate that, MVRD as a multi-view motion benchmark, can improve the performance and generalization of view variations.

\noindent\textbf{Single-View Performance.} Table~\ref{tab:movement} summarizes the quantitative results of single-view baselines and our method under the multi-view movement scenario, plus stationary and speaking scenarios at the center view. Traditional methods (ICA~\cite{ICA}, CHROM~\cite{CHROM}, POS~\cite{POS})
 exhibit good performance only in stationary conditions but degrade significantly in dynamic settings due to motion artifacts. Supervised learning based models (PhysNet~\cite{ST-networks}, PhysFormer~\cite{physformer}, iBVPNet~\cite{iBVPNet}) demonstrate robustness, with PhysFormer outperforming PhysNet and iBVPNet under motion. Contrast-Phys+~\cite{contrast-phys+} achieves competitive accuracy in stationary scenarios but struggles with motion.
 
\noindent\textbf{Multi-View Fusion Performance.} Our method leverages complementary cross-view cues and noise-aware fusion to achieve superior performance. As shown in Table~\ref{tab:view-missing}, under the movement scenario, it attains an MAE of 0.90 bpm and Pearson R of 0.99, substantially outperforming single-view baselines and unsupervised approaches.

\noindent\textbf{View-Missing Robustness.}
In Table~\ref{tab:view-missing}, to simulate practical deployment scenarios with missing camera views, we evaluate the fusion model~\cite{multiview} under view-missing settings that are applied consistently during both training and inference. The missing view is implemented by setting the corresponding video input to zero. The absence of the center view leads to the significant performance drop, indicating its dominant role in motion-aware fusion. Nevertheless, MVRD-rPPG shows the efficiency of cross-view feature compensation.

\noindent\textbf{Cross-Dataset Generalization.} 
To evaluate the generalization capability of our method, we compare single-view baselines with our approach on the unseen datasets. All models are trained only on the movement subset of the MVRD dataset. We conduct cross-dataset testing on PURE~\cite{PURE} and UBFC-rPPG~\cite{UBFC-rPPG} without any fine-tuning. As reported in Table~\ref{tab:cross_dataset}, despite the substantial differences in recording conditions and acquisition devices between MVRD and testing datasets, our method achieves the lowest MAE and highest R on two datasets, demonstrating excellent cross-domain generalization ability.

Overall, the above experiments show that the MVRD dataset we constructed fills the gap in the existing rPPG tasks that lack multi-view and motion interference scenarios, and can effectively explore the complementarity of multi-view features. Further, our MVRD-rPPG improves the accuracy of physiological signals by using multi-view compensation mechanism in complex dynamic scenes.


\subsection{Ablation Study}
To investigate the contribution of each component and training setup, we conduct ablation experiments from two aspects: the components of the model structure and the loss function.

\noindent\textbf{Contribution of Structural Modules.} We evaluate the performance of the core components of the model to analyze their individual roles and advantages, including Adaptive Temporal Optical Compensation (ATOC) and Multi-View Correlation-Aware Attention (MVCA). Table~\ref{tab:component_ablation} shows that the model fails to suppress motion artifacts when either component is removed (denoted as "w/o", i.e., without). The best result is obtained by using both ATOC and MVCA.

\begin{table}[t]
\setlength{\tabcolsep}{10.4pt} 
\centering
\scriptsize  
\caption{Ablation study on model structural components in MVRD Movement. Bold fonts denote the best performance.}
\label{tab:component_ablation}
\begin{tabular}{lccc}
\toprule
Scene & MAE & RMSE& R \\
\midrule
w/o ATOC+MVCA & 5.12 & 6.67 & 0.78 \\
w/o ATOC      & 4.23 & 5.41 & 0.82 \\
w/o MVCA      & 2.05  & 3.45  & 0.92 \\
\textbf{Our Method} & \textbf{0.90} & \textbf{1.02} & \textbf{0.99} \\
\bottomrule
\end{tabular}
\end{table}

\noindent\textbf{Effect of Loss Function Design.} We further investigate the impact of different loss functions. Our MVRD-rPPG method can encourage the generation of realistic rPPG signals by strategically combining a Pearson-based correlation loss $\mathcal{L}_{\text{Pearson}}$, PSD consistency loss $\mathcal{L}_{\text{PSD}}$ and adversarial loss $\mathcal{L}_{G}$, where the three loss functions complement each other to address the key challenges in rPPG signal estimation. As shown in Table~\ref{tab:loss_ablation}, the results indicate that each loss function contributes positively, and their combination yields the best performance. The correlation loss and PSD consistency loss stabilizes temporal signal learning, whereas the adversarial loss improves signal realism and reduces motion noise.

\begin{table}[t]
\setlength{\tabcolsep}{9.5pt}
\centering
\scriptsize  
\caption{Ablation study on loss terms on MVRD Movement.}
\label{tab:loss_ablation}
\begin{tabular}{lccc}
\toprule
Training & MAE & RMSE & R \\
\midrule
w/o $\mathcal{L}_{\text{Pearson}}$       & 1.83 & 2.09 & 0.96 \\
w/o $\mathcal{L}_{\text{PSD}}$       & 1.52 & 1.78 & 0.98 \\
w/o $\mathcal{L}_{G}$   & 1.10 & 1.72 & 0.98 \\
\textbf{Total Loss}    & \textbf{0.90} & \textbf{1.02} & \textbf{0.99} \\
\bottomrule
\end{tabular}
\end{table}

In general, MVRD dataset integrates physiological data from different views, forcing the model to learn general physiological features that are independent of the view variations. Further, our MVRD-rPPG method solves the shortcomings of existing rPPG models in terms of generalization under view switching and motion interference scenarios.

\section{Conclusion}
In this paper, we present a new MVRD-bench for dynamic multi-view remote photoplethysmography under occlusion. Specifically, we construct the MVRD dataset with synchronized recordings from three viewpoints under stationary, speaking, and head-movement scenarios, and develop a multi-view rPPG learning framework for robust rPPG estimation by introducing motion compensation and adaptive cross-view fusion. Extensive experiments on MVRD and public datasets demonstrate the effectiveness and generalization ability of the proposed approach. We hope that MVRD-Bench can provide a useful benchmark for future research on robust multi-view rPPG in real-world scenarios. In future work, we further extend our MVRD-rPPG framework to real-world applications like driving, sleeping, or working.

\bibliography{main}

\end{document}